\documentclass[10pt,twocolumn,letterpaper]{article}

\usepackage[accsupp]{axessibility}        
\usepackage[pagenumbers]{cvpr} 

\usepackage{graphicx}
\usepackage{amsmath}
\usepackage{amssymb}
\usepackage{booktabs}
\usepackage{multirow}
\usepackage{threeparttable}
\usepackage[pagebackref,breaklinks,colorlinks]{hyperref}

\usepackage[capitalize]{cleveref}
\crefname{section}{Sec.}{Secs.}
\Crefname{section}{Section}{Sections}
\Crefname{table}{Table}{Tables}
\crefname{table}{Tab.}{Tabs.}

\begin{document}

\title{STRPM: A Spatiotemporal Residual Predictive Model for High-Resolution Video Prediction}
\author{Zheng Chang$^{1,2}$,~Xinfeng Zhang$^{1}$,{~Shanshe Wang$^{3}$\thanks{Corresponding author: Shanshe Wang, sswang@pku.edu.cn.
This work was supported in part by the National Natural Science Foundation of China (62025101, 62072008, 62071449, U20A20184), National Key Research and Development Project of China (2019YFF0302703, 2021YFF0900503) and High-performance Computing Platform of Peking University, which are gratefully acknowledged.
}},~Siwei Ma$^{3}$,~and Wen Gao$^{1,3}$\\
$^{1}$School of Computer Science and Technology, \\University of Chinese Academy of Sciences, Beijing, China\\
$^{2}$Institute of Computing Technology, Chinese Academy of Sciences, Beijing, China\\
$^{3}$National Engineering Research Center of Visual Technology,\\
School of Computer Science, Peking University, Beijing, China\\
{\tt\small changzheng18@mails.ucas.ac.cn, xfzhang@ucas.ac.cn, \{sswang, swma, wgao\}@pku.edu.cn}}

\maketitle

\begin{abstract}
   Although many video prediction methods have obtained good performance in low-resolution (64$\sim$128) videos, predictive models for high-resolution (512$\sim$4K) videos have not been fully explored yet, which are more meaningful due to the increasing demand for high-quality videos. Compared with low-resolution videos, high-resolution videos contain richer appearance (spatial) information and more complex motion (temporal) information. In this paper, we propose a Spatiotemporal Residual Predictive Model (STRPM) for high-resolution video prediction. On the one hand, we propose a Spatiotemporal Encoding-Decoding Scheme to preserve more spatiotemporal information for high-resolution videos. In this way, the appearance details for each frame can be greatly preserved. On the other hand, we design a Residual Predictive Memory (RPM) which focuses on modeling the spatiotemporal residual features (STRF) between previous and future frames instead of the whole frame, which can greatly help capture the complex motion information in high-resolution videos. In addition, the proposed RPM can supervise the spatial encoder and temporal encoder to extract different features in the spatial domain and the temporal domain, respectively. Moreover, the proposed model is trained using generative adversarial networks (GANs) with a learned perceptual loss (LP-loss) to improve the perceptual quality of the predictions. Experimental results show that STRPM can generate more satisfactory results compared with various existing methods.
\end{abstract}

\section{Introduction}\label{intro}
Video prediction is a key component of representation learning due to its great ability in modeling meaningful representations for natural videos and has been applied to various video processing applications, such as video coding \cite{ma2019image}, precipitation nowcasting \cite{xingjian2015convolutional}, robotic control \cite{finn2016unsupervised}, autonomous driving \cite{bhattacharyya2018long} and so on.
Different from video interpolation \cite{niklaus2018context,meyer2018phasenet}, video prediction (extrapolation) is more challenging by merely utilizing limited information from previous frames to predict the unknown future frames.
Motivated by the advantages of deep learning technologies in extracting deep features, in recent years, various learning-based methods have been proposed for video prediction
which can be summarized into three types.

\begin{figure}[t]
  \centering
  \includegraphics[width=\columnwidth]{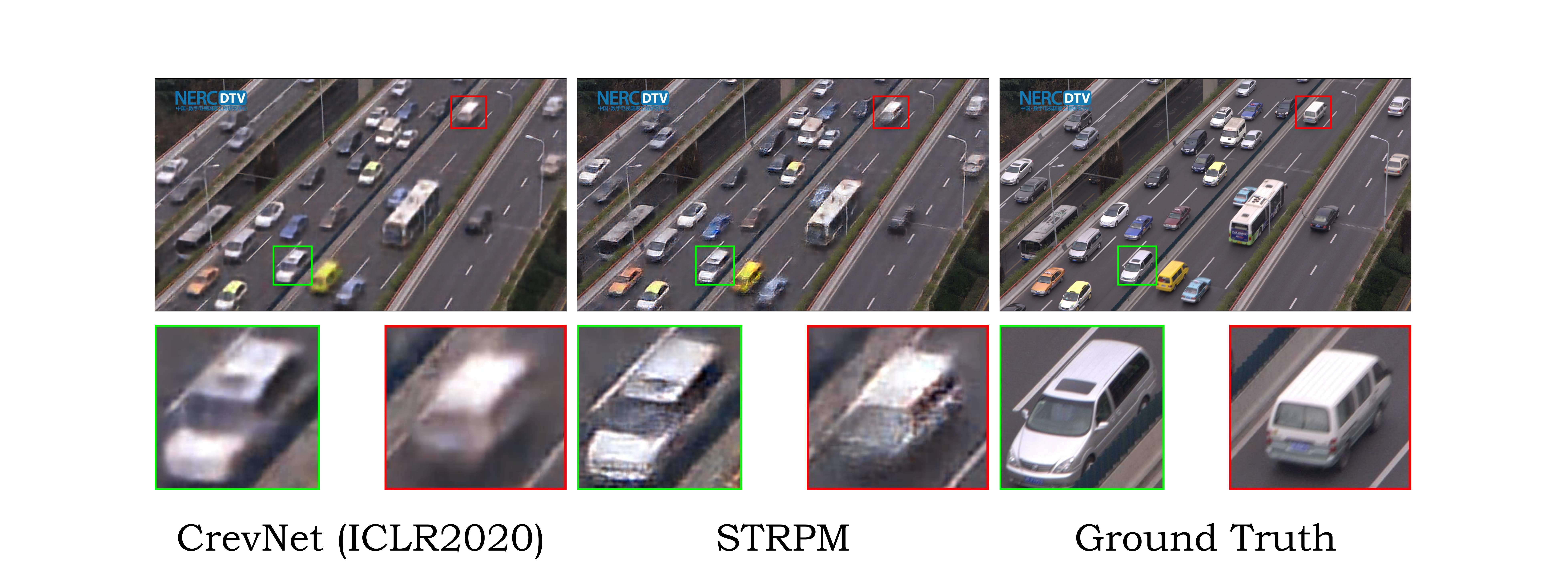}
  \caption{Qualitative results between the proposed STRPM and the state-of-the-art method CrevNet \cite{yu2019efficient} on the SJTU4K dataset (4K: 2160$\times$3840 resolution, 4 frames $\rightarrow$ 1 frame). STRPM has generated much better visual details compared with CrevNet. }\label{fig:sjtu4k01}
\end{figure}
The first type of methods \cite{xingjian2015convolutional,srivastava2015unsupervised,wang2017predrnn,wang2019eidetic,yu2019efficient,wu2021motionrnn} utilize Recurrent neural networks (RNN) to progressively predict video frames due to their unique advantages in sequence learning and have obtained some satisfactory results. However, the predictions from RNN-based methods are typically blurry due to the standard mean square error based loss function.
To solve this problem, the second type of methods \cite{babaeizadeh2018stochastic,denton2018stochastic,villegas2019high,franceschi2020stochastic,xu2020video} utilize deep stochastic models to predict different futures instead of an averaged future for different samples and the third type of methods \cite{goodfellow2014generative,mathieu2016deep,lee2018stochastic,kwon2019predicting,chen2020long} employ generative adversarial networks (GANs) \cite{goodfellow2014generative} and additional perceptual loss functions to augment the visual qualities of the predictions.

Although the above methods have obtained some satisfactory results,
the resolutions of video datasets utilized in the above methods are usually low (64$\sim$128), and the performance in high-resolution (512$\sim$4K) videos is still hardly satisfactory (shown in Figure \ref{fig:sjtu4k01}),
preventing their adaptability and practicability into real scenarios. There are mainly two challenges restricting the resolution of predictions. The first challenge is that high-resolution videos usually contain more complex visual details. However, limited by the computation resources, videos are usually encoded to low-dimensional features and then decoded back to the video frames, during which, lots of visual details can be abandoned. The second challenge is that the motion information in high-resolution videos typically involves multiple objects, which is much more complex and hard for traditional predictive memory to predict. To deal with the above two problems in high-resolution video prediction, the appearance information in the spatial domain and the motion information in the temporal domain need to be carefully reconsidered.

In this paper, we propose a Spatiotemporal Residual Predictive Model (STRPM) to deal with the above two challenges.
Firstly, to predict more satisfactory appearance details for each frame, we novelly propose the spatiotemporal encoding-decoding scheme, which utilizes independent encoders to extract deep features in both spatial and temporal domains. In this way, both the spatial and temporal information can no longer affect each other and more visual details can be preserved.
Secondly, to accurately model the complex motion information in high-resolution videos, we designed a Residual Predictive Memory (RPM) to focus on modeling the inter-frame spatiotemporal residual features (STRF) with a relatively low computation load and fewer parameters.
Moreover, since the encoded spatial and temporal features will be fed into the spatial and temporal modules in RPM for transitions in vertical (spatial domain) and horizontal (temporal domain) directions, the RPM can indirectly supervise the spatial encoder and the temporal encoder to extract corresponding features in the spatial domain and the temporal domain.

By jointly using the encoded spatiotemporal features and STRF, more reliable spatiotemporal features for future frames can be predicted, which will be further decoded back to the high-dimensional data space with the help of spatiotemporal decoders.
Furthermore, in the training stage, the standard MSE loss, the adversarial loss as well as the learned perceptual loss are jointly utilized to improve the visual quality of the predictions. Experimental results show that the proposed model can achieve state-of-the-art performance compared with other methods.

\section{Related Work}
In recent years, many learning-based predictive models have been applied in video prediction. \cite{ranzato2014video} first utilized language modeling for video prediction, which was further improved by \cite{srivastava2015unsupervised} using Long Short-Term Memories (LSTMs) \cite{hochreiter1997long}, denoted as FC-LSTM. To improve the model perception to visual data, \cite{xingjian2015convolutional} integrated convolutional operations to FC-LSTMs (ConvLSTM) and achieved significant improvements on the Moving MNIST dataset.

However, the above works only focus on the inter-frame temporal information (motion information) and ignored intra-frame spatial information (appearance information). To preserve the appearance information for videos, \cite{wang2017predrnn} designed an appearance-preserving block for ConvLSTM (PredRNN).
\cite{wang2018predrnn++} further improved PredRNN by solving the gradient propagation difficulties in deep predictive models (PredRNN++) and integrating 3D convolution operations and RECALL gate to enhance the ability to capture both long-term and short-term dependencies of the predictive model (E3D-LSTM). To further improve the visual quality of the predictions, \cite{yu2019efficient} proposed a conditionally reversible network (CrevNet) to preserve the spatiotemporal information for the inputs and \cite{jin2020exploring} leveraged the high-frequency information to preserve visual details for videos.

However, the above works can only generate an averaged future for all samples due to the standard MSE loss function and the predictions are usually blurry. To solve this problem, a variety of methods have been proposed.
On the one hand, some methods aim to predict different futures for different samples.
\cite{babaeizadeh2018stochastic} proposed a stochastic variational video prediction (SV2P) method to predict a different possible future for each sample based on the latent variables.
\cite{denton2018stochastic,xu2020video} proposed video generation models with a learned prior over stochastic latent variables for video prediction.
\cite{franceschi2020stochastic} proposed a stochastic temporal model for video prediction whose dynamics are governed in a latent space by a residual update rule.

\begin{figure*}[t]
  \centering
  \includegraphics[width=\textwidth]{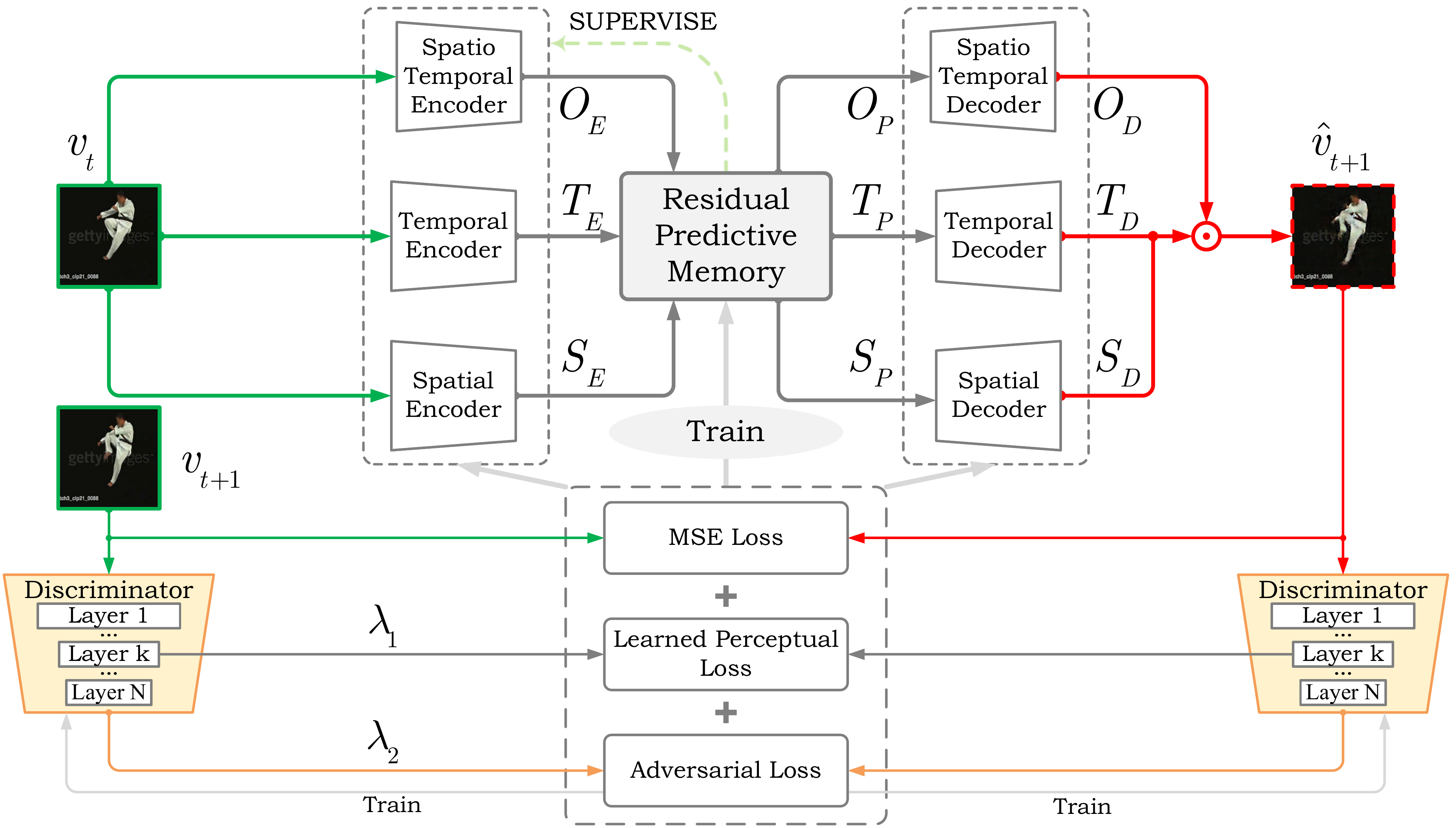}
  \caption{The structure of the proposed spatiotemporal residual predictive model (STRPM). The green arrows denote the input information flows and the red arrows denote the predicted information flows.}\label{fig:model}
\end{figure*}
On the other hand, some works aim to improve the standard MSE-based loss function.
\cite{mathieu2016deep} proposed three different and complementary feature learning strategies to predict naturalistic videos. Besides, motivated by the great power in generating naturalistic images, generative adversarial networks (GANs) were employed by \cite{lee2018stochastic} to generate realistic results and \cite{kwon2019predicting} utilized CycleGAN \cite{zhu2017unpaired} to further improve the perceptual quality of the predictions.
Although some improvements have been achieved in above works and the resolutions of the predicted videos have been improved in \cite{yu2019efficient,chang2021stae} (256$\sim$512), the unacceptable computation load and quality distortions prevent them from predicting videos with higher resolution (512$\sim$4K). To solve the above problems, we propose a spatiotemporal residual predictive model (STRPM) for high-resolution video prediction with an acceptable computation load. Moreover, by using the proposed learned perceptual loss, more naturalistic videos can be generated from the proposed method.

\section{The Spatiotemporal Residual Predictive Model}
In this section, we introduce the proposed Spatiotemporal Residual Predictive Model (STRPM) in detail. The overall structure of the proposed model is shown in Figure \ref{fig:model}. Different from low-resolution videos, high-resolution videos contain more complex texture details and more variable motion information, motivated by which, two problems for high-resolution video prediction urgently need to be solved:
\begin{itemize}
  \item How to preserve more visual details for each frame?
  \item How to predict more accurate motion information between frames?
\end{itemize}
We propose the Spatiotemporal Residual Predictive Model (STRPM) to solve the above problems.
\begin{figure*}[t]
  \centering
  \includegraphics[width=\textwidth]{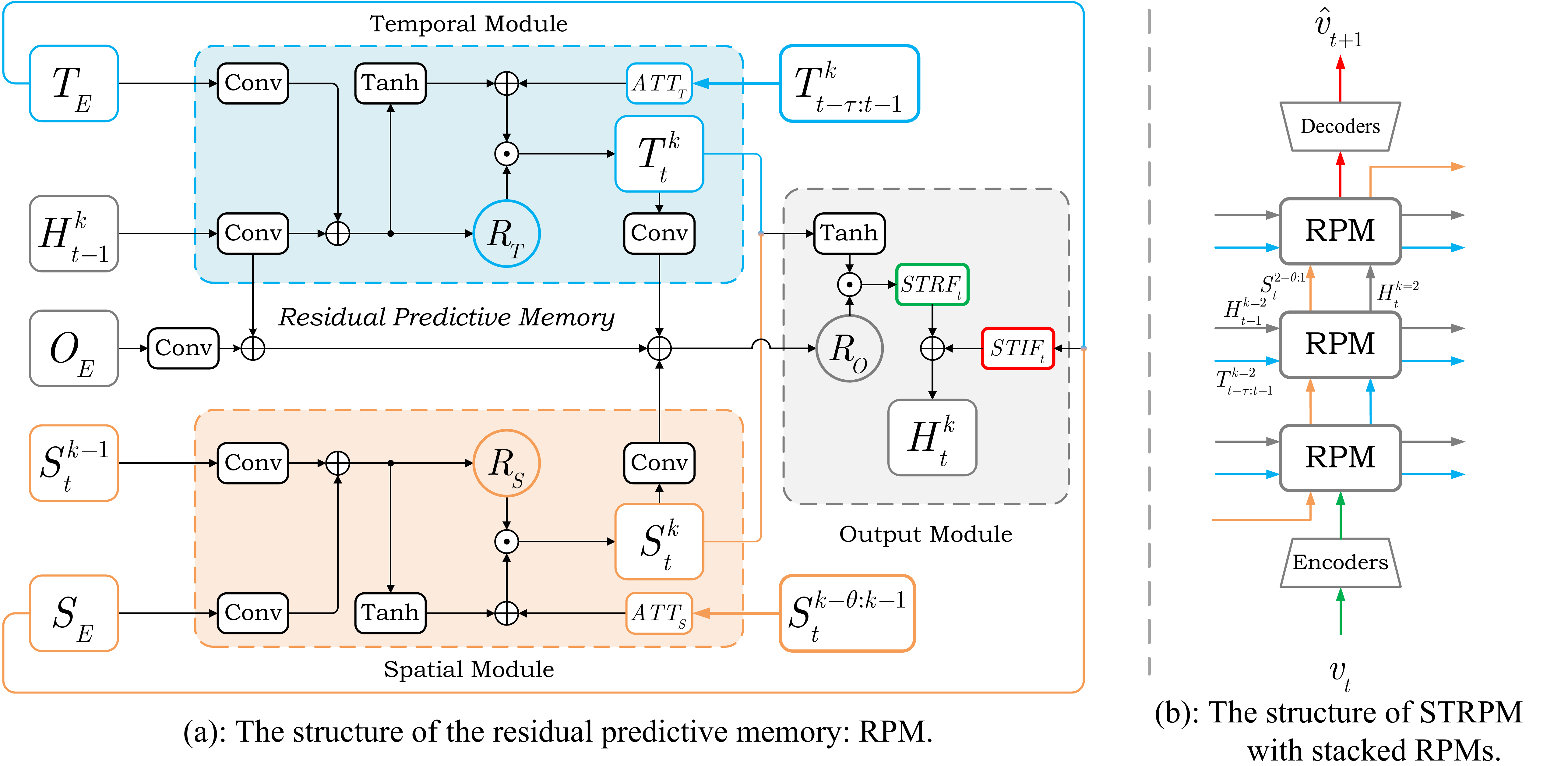}
  \caption{The structure of the proposed residual predictive memory: RPM. The temporal module and the spatial module can indirectly supervise the temporal encoder and the spatial encoder to extract different features in the temporal domain and the spatial domain.}\label{fig:memory}
\end{figure*}
\subsection{Spatiotemporal Encoding-Decoding Scheme}\label{STRPM}
To reduce the computation resources, video frames are typically encoded to low-dimensional features using a single encoder in video prediction. However, the temporal information and the spatial information will affect each other and the predictive memories have to further extract the temporal and spatial information to predict future frames, during which, lots of spatiotemporal information may be lost, making it very difficult to reconstruct satisfactory visual details for each frame.
To solve this problem (the first problem), we novelly utilize multiple spatiotemporal encoders to independently extract deep features in both temporal and spatial domains. In this way, the spatial information and the temporal information will no longer affect each other, making it easier for the predictive memories to utilize the spatiotemporal information for video prediction. The encoding process can be expressed as follows,
\begin{equation}
  (T_E,S_E,O_E) = (Enc_T(v_t),Enc_S(v_t),Enc_O(v_t)),
\end{equation}
where $v_t$ denotes the $t^{th}$ frame in source video $V$. $Enc_T(\cdot),Enc_S(\cdot),Enc_O(\cdot)$ denote the temporal, spatial and spatiotemporal encoders respectively. $T_E,S_E,O_E$ denote the encoded low-dimensional temporal, spatial and spatiotemporal features at time step $t$, respectively.

In particular, the above encoded features $T_E,S_E,O_E$ will be fed into the corresponding modules in the proposed residual predictive memory: RPM, which will be detailedly introduced in section \ref{RPM}. In this way, the RPM can indirectly supervise different encoders to extract different features in different domains. And the predicted spatiotemporal features can be represented as follows,
\begin{equation}
  (T_P,S_P,O_P) = RPM(T_E,S_E,O_E, \mathcal{T}, \mathcal{S}),
\end{equation}
where $T_P,S_P,O_P$ denote the predicted temporal, spatial and spatiotemporal features at time step $t$ from RPM , respectively. $\mathcal{T}, \mathcal{S}$ are the preserved temporal and spatial information.

Similar as the encoding process, to decode more spatiotemporal details, we also utilize multiple spatiotemporal decoders to decode the predicted features from low-dimensional feature space back to high-dimensional temporal and spatial data space respectively, which can be expressed as follows,
\begin{equation}
  (T_D,S_D,O_D) = (Dec_T(T_P),Dec_S(S_P),Dec_O(O_P)),
\end{equation}
where $Dec_T(\cdot),Dec_S(\cdot),Dec_O(\cdot)$ denote the temporal, spatial and spatiotemporal decoders, respectively. $T_D,S_D,O_D$ denote the decoded high-dimensional temporal, spatial and spatiotemporal features, respectively.

By jointly utilizing the decoded high-dimensional features, the predicted frame at time step $t$ can be represented as follows,
\begin{equation}
  \hat{v}_{t+1} = O_D\odot\tanh(W_{1\times1}\ast[T_D,S_D]),
\end{equation}
where $\hat{v}_{t+1}$ denotes the predicted frame at time step $t$, $\odot,\ast$ denote the hadamard product and convolutional operators.

\subsection{The Residual Predictive Memory: RPM}\label{RPM}
Current predictive memories aim to predict future frames by learning a single representation for the whole frame, containing both the spatial representation and the temporal representation, which is efficient to predict simple motions. However, high-resolution videos usually contained much more complex motion information compared with low-dimensional videos. To deal with this special characteristic in high-resolution videos (the second problem), we design a Residual Predictive Memory (RPM) to focus on modeling inter-frame motion information by predicting the spatiotemporal residual features (STRF) between previous and future frames in the feature space,
which is shown in Figure \ref{fig:memory}(a). In addition, compared with traditional ST-LSTM structure \cite{wang2017predrnn}, the proposed RPM also benefits from more efficient state-to-state transitions (fewer gates) and wider spatiotemporal receptive field (simultaneously utilizing multiple spatiotemporal states).
To further extract more efficient deep spatiotemporal features, multiple RPMs are typically stacked into a single model, as shown in Figure \ref{fig:memory}(b). For RPM at time step $t$ in layer $k$, the encoded features $T_E, S_E, O_E$ are fed into the corresponding modules of RPM. In this way, the proposed RPM can indirectly supervise the spatial encoder and the temporal encoder to extract different deep features in the spatial domain and the temporal domain. In particular, for $k>1$, the encoded features are represented with the hidden state from the previous layer, i.e., $T_E, S_E, O_E = H_t^{k-1}$.

For each RPM, there are seven inputs: $T_E$, the encoded features for the temporal module; $S_E$, the encoded features for the spatial module; $O_E$, the encoded spatiotemporal features for the output module; $H_{t-1}^k$, the hidden state from previous time step; $S_t^{k-1}$, the previous spatial state from previous layer $k-1$; $\mathcal{T}:T_{t-\tau:t-1}^k$, the previous $\tau$ temporal states; $\mathcal{S}:S_t^{k-\theta:k-1}$, the previous $\theta$ spatial states. To further improve local perception to videos, the input states are typically preprocessed using convolutional layers:
\begin{eqnarray}
  (TF_t^k, SF_t^k, OF_t^k) &=& (W_t\ast T_E,W_s\ast S_E,W_o\ast O_E),\nonumber \\
  (HF_t^k,MF_t^k) &=& (W_h\ast H_{t-1}^k,W_m\ast S_t^{k-1}),
\end{eqnarray}
where $W$ denotes the parameters of the integrated convolutional layers. $TF_t^k$, $SF_t^k$, $OF_t^k$, $HF_t^k$, $MF_t^k$ denote the extracted deep features from $T_E$, $S_E$, $O_E$, $H_{t-1}^k$, $S_t^{k-1}$, respectively. Then the extracted features will be fed into RPM at time step $t$ in layer $k$.

For both the temporal and spatial module, two residual gates are designed to model the inter-frame residual information, which is shown as follows,
\begin{align}
  R_T &= \sigma(TF_t^k + HF_t^k),\nonumber\\
  R_S &= \sigma(SF_t^k + MF_t^k),
\end{align}
where $R_T,R_S$ denote the temporal and spatial residual gates, respectively.

As shown in Figure \ref{fig:memory}(a), the temporal module (blue block) of RPM is utilized to capture reliable motion information between frames. To preserve more useful temporal information from the past, RPM jointly utilizes multiple temporal states, and the transitions can be expressed as follows,
\begin{eqnarray}
  T_t^k &=& R_T\odot(\tanh(TF_t^k + HF_t^k)+ATT_T(\mathcal{T})).
\end{eqnarray}
The predicted temporal residual state $T_t^k$ consists of two terms, where the first term $R_T\odot \tanh(TF_t^k + HF_t^k)$ represents the encoded features from current input and the second term $R_T\odot ATT_t(\mathcal{T})$ represents the preserved temporal information from previous $\tau$ time steps. In this way, more useful temporal information can be kept from a longer past. In particular, $ATT_T(\cdot)$ denotes the temporal attention network which is constructed with convolutional layers and can help merge the multiple temporal states to a single one.

In the spatial module (orange block), by utilizing the multiple spatial states $\mathcal{S}:S_t^{k-\theta:k-1}$, both low-level texture information and high-level semantic information can be jointly utilized, and similar to the temporal module, the state-to-state transitions can be represented as follows,
\begin{eqnarray}
  S_t^k &=& R_S\odot(\tanh(SF_t^k + MF_t^k)+ATT_S(\mathcal{S})),
\end{eqnarray}
where $S_t^k$ denotes the predicted spatial residual state and $ATT_S$ denotes the spatial attention network.

The predicted temporal residual state $T_t^k$ and the predicted spatial residual state $S_t^k$ will be further aggregated to the final hidden state in the output module (gray block):
\begin{eqnarray}\label{equ:RPM_information_fusion}
  R_O &=& \sigma(OF_t^k + HF_t^k + W_{os}\ast S_t^k + W_{ot}\ast T_t^k),\nonumber\\
  STRF_t &=& R_O\odot\tanh(W_{1\times1}\ast[T_t^k,S_t^k]),\nonumber\\
  STIF_t &=& W_{1\times1}\ast[T_E,S_E],\nonumber\\
  H_t^k &=& STIF_t^k + STRF_t^k,
\end{eqnarray}
where $R_O$ denotes the output residual gate, which is utilized to aggregate the predicted temporal and spatial residual information.
$H_t^k$ denotes the final hidden state.
In particular, the hidden state $H_t^k$ consists of two terms, where the first term $STIF_t^k$ denotes the spatiotemporal input features and the second term $STRF_t^k$ denotes the predicted spatiotemporal residual features between previous and future frames.

\subsection{Training Details}
In training stage, to predict more naturalistic results, the proposed model is trained with the help of GANs and the whole model consists of two submodules: predictor $P$ which is utilized to generate future frames and discriminator $D$ which is utilized to judge whether the input frames are real or generated. The adversarial loss for both modules can be expressed as follows,
\begin{eqnarray}
  \mathcal{L}_{GAN}(D) &=& -\sum_{t=2}^{T}[\log(D(v_t)) + \log(1-D(\hat{v}_t))],\nonumber\\
  \mathcal{L}_{GAN}(P) &=& -\sum_{t=2}^{T}[\log(D(\hat{v}_t)),
\end{eqnarray}
where $T$ denotes total number of the time steps. $v$ and $\hat{v}$ denote the input and predicted frames respectively.

Since the discriminators in GANs can model the distribution of the input data (fake or real), we utilize the feature map from the layer $k$ of the discriminator as the learned perceptual representations for current input. And a learned perceptual loss, which can indicate the perceptual distribution of the inputs, is represented as follows (Figure \ref{fig:model}),
\begin{equation}
  \mathcal{L}_{LP} = \sum_{t=2}^{T}\mathcal{L}_2[D_k(v_t),D_k(\hat{v}_{t})],
\end{equation}
where $D_k$ denotes the $k^{th}$ layer of the discriminator $D$ (the bottom layer in our method). $\mathcal{L}_2(\cdot)$ denotes the standard MSE loss function.
By using the additional loss functions, more naturalistic results can be predicted and the final loss function for the predictor can be expressed as follows,
\begin{equation}
  \mathcal{L}_P = \mathcal{L}_{MSE} + \lambda_1\mathcal{L}_{LP} + \lambda_2\mathcal{L}_{GAN}(P),
\end{equation}
where $\lambda_1,\lambda_2$ control the relative importance.

\section{Experiments}\label{experiment}
In this section, we evaluate all models on three high-resolution datasets, UCF Sports dataset ($480\times 720$) \cite{rodriguez2008action}, Human3.6M dataset ($1000\times 1000$) \cite{h36m_pami} and SJTU4K dataset ($2160\times 3840$) \cite{song2013sjtu}. We stack 16 RPMs to the proposed STRPM and the integrated convolutional operations are set with a kernel size $5\times 5$. The stride is set to 1 for each
dimension. We set the number of previous spatiotemporal states $\tau$, $\theta$ to 5. The hidden states for STRPM and the discriminator are set with 128 channels. All models are implemented using Pytorch and trained with Adam optimizer. In the training stage, models are trained to predict the next frame with 4 successive frames as the input on all datasets. In the testing stage, models are evaluated to predict multiple frames. The balance weights $\lambda_1$, $\lambda_2$ are set to 0.01, 0.001 for UCF Sports and Human3.6M datasets, and 0.005, 0.0005 for SJTU4K dataset.
\begin{figure}[t]
  \centering
  \includegraphics[width=\columnwidth]{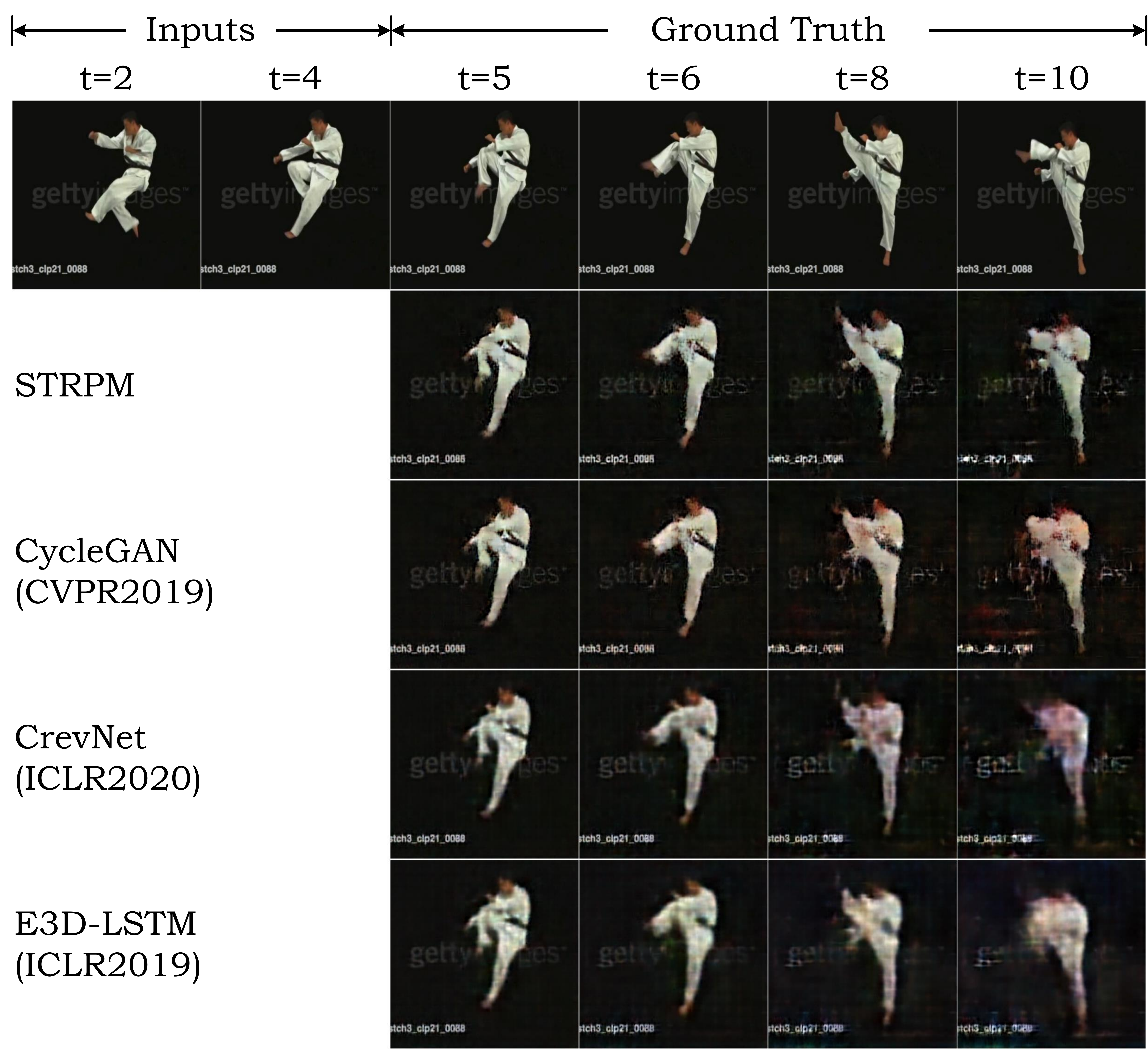}
  \caption{The generated examples on the UCF Sports test set (4 frames $ \rightarrow$ 6 frames).}\label{fig:ucfsport}
\end{figure}
\subsection{UCF Sports Dataset}
\begin{table*}[!htb]
  \centering
  \caption{Quantitative results of different methods on the UCF Sports (4 frames $ \rightarrow$ 6 frames) and Human3.6M (4 frames $ \rightarrow$ 4 frames) datasets. Lower LPIPS($10^{-2}$) and higher PSNR(dB) scores indicate better results.}
  \label{tab:ucf_human36m}
   { \begin{tabular}{lcccc}
    \toprule
    \multirow{3}{*}{Method}&\multicolumn{2}{c}{UCF Sports}&\multicolumn{2}{c}{Human3.6M}\cr
    \cmidrule(lr){2-3}\cmidrule(lr){4-5}
    &$t=5$&$t=10$&$t=5$&$t=8$\cr
    \cmidrule(lr){2-2} \cmidrule(lr){3-3}\cmidrule(lr){4-4} \cmidrule(lr){5-5}
    &PSNR$\uparrow$ / LPIPS$\downarrow$&PSNR$\uparrow$ / LPIPS$\downarrow$&PSNR$\uparrow$ / LPIPS$\downarrow$&PSNR$\uparrow$ / LPIPS$\downarrow$\cr
    \midrule
    BeyondMSE (ICLR2016) \cite{mathieu2016deep}                         &26.42 / 29.01  &18.46 / 55.28   &-&-\cr
    PredRNN (NeurIPS2017) \cite{wang2017predrnn}                        &27.17 / 28.15  &19.65 / 55.34   &31.91 / 12.62  &25.65 / 14.01\cr
    PredRNN++ (ICML2018) \cite{wang2018predrnn++}                       &27.26 / 26.80  &19.67 / 56.79   &32.05 / 13.85  &27.51 / 14.94\cr
    SAVP (arXiv 2018) \cite{lee2018stochastic}                          &27.35 / 25.45  &19.90 / 49.91   &-&-\cr
    SV2P (ICLR2018) \cite{babaeizadeh2018stochastic}                    &27.44 / 25.89  &19.97 / 51.33   &31.93 / 13.91  &27.33 / 15.02\cr
    HFVP (NeurIPS2019) \cite{villegas2019high}                          &-              &-               &32.11 / 13.41  &27.31 / 14.55\cr
    E3D-LSTM (ICLR2019) \cite{wang2019eidetic}                          &27.98 / 25.13  &20.33 / 47.76   &32.35 / 13.12  &27.66 / 13.95\cr
    CycleGAN (CVPR2019) \cite{kwon2019predicting}                       &27.99 / 22.95  &19.99 / 44.93   &32.83 / 10.18  &28.26 / 11.03\cr
    CrevNet (ICLR2020) \cite{yu2019efficient}                           &28.23 / 23.87  &20.33 / 48.15   &33.18 / 11.54  &28.31 / 12.37\cr
    MotionRNN (CVPR2021) \cite{wu2021motionrnn}                                                &27.67 / 24.23  &20.01 / 49.20   &32.20 / 12.11  &28.03 / 13.29\cr
    \midrule
    STRPM                                                   &\textbf{28.54} / \textbf{20.69}  &\textbf{20.59} / \textbf{41.11}    &\textbf{33.32} / \textbf{9.74}  &\textbf{29.01} / \textbf{10.44}\cr
    \bottomrule
    \end{tabular}}
\end{table*}
The UCF Sports dataset contains a series of human actions collected from various sports events and are typically captured on broadcast television channels such as the BBC and ESPN. A total of 150 videos with resolution of $480\times 720$ are contained in the UCF Sports dataset. We resize each frame to $512\times 512$. 6,288 sequences are for training and 752 for testing.
Figure \ref{fig:ucfsport} shows the qualitative results generated from different methods, where the proposed method obviously outperforms others with more naturalistic predictions. In Table \ref{tab:ucf_human36m}, we utilize the Peak Signal to Noise Ratio (PSNR) to represent the objective quality and the Learned Perceptual Image Patch Similarity (LPIPS) \cite{zhang2018unreasonable} to represent the perceptual quality. The quantitative results show that the proposed method achieves the best PSNR score and LPIPS scores.

\begin{figure}[t]
  \centering
  \includegraphics[width=\columnwidth]{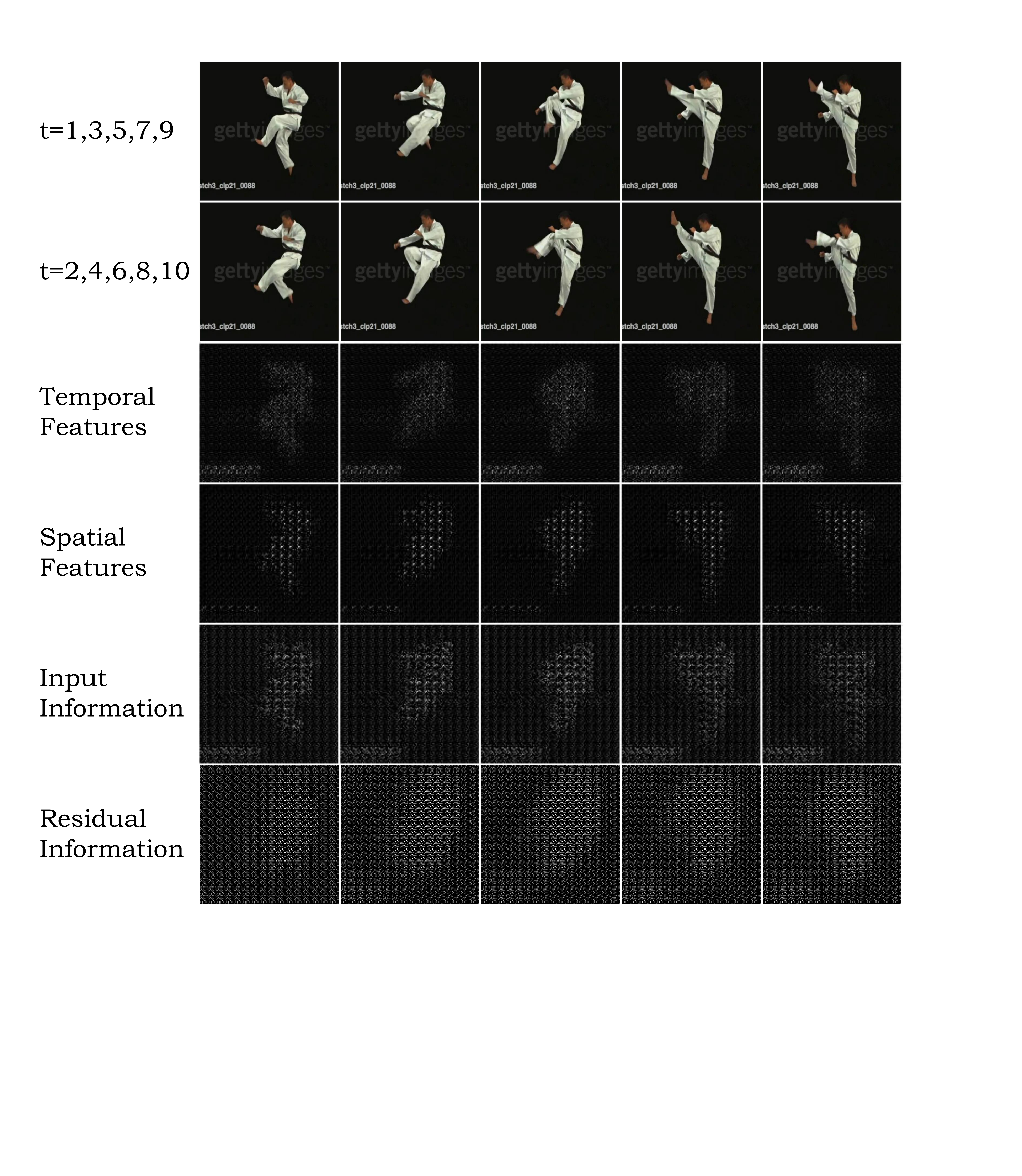}
  \caption{The visualization results of the spatiotemporal encoding scheme and the residual predictive memory. \textbf{Temporal Features} and \textbf{Spatial Features} denote the encoded features from the temporal encoder and the spatial encoder. \textbf{Input Information} and \textbf{Residual Information} denote $STIF$ and $STRF$ in Equation \ref{equ:RPM_information_fusion}.}\label{fig:features}
\end{figure}
\begin{figure*}[!htb]
  \centering
  \includegraphics[width=\textwidth]{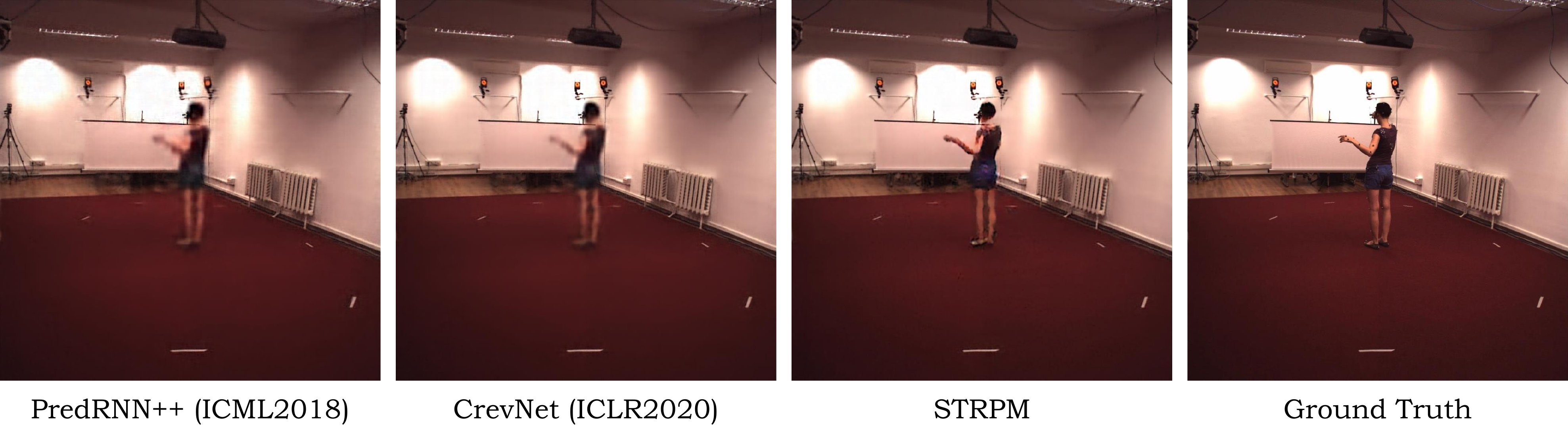}
  \caption{The generated examples on the Human3.6M dataset (4 frames $ \rightarrow$ 1 frame).}\label{fig:human36m}
\end{figure*}
\begin{figure*}[!htb]
  \centering
  \includegraphics[width=\textwidth]{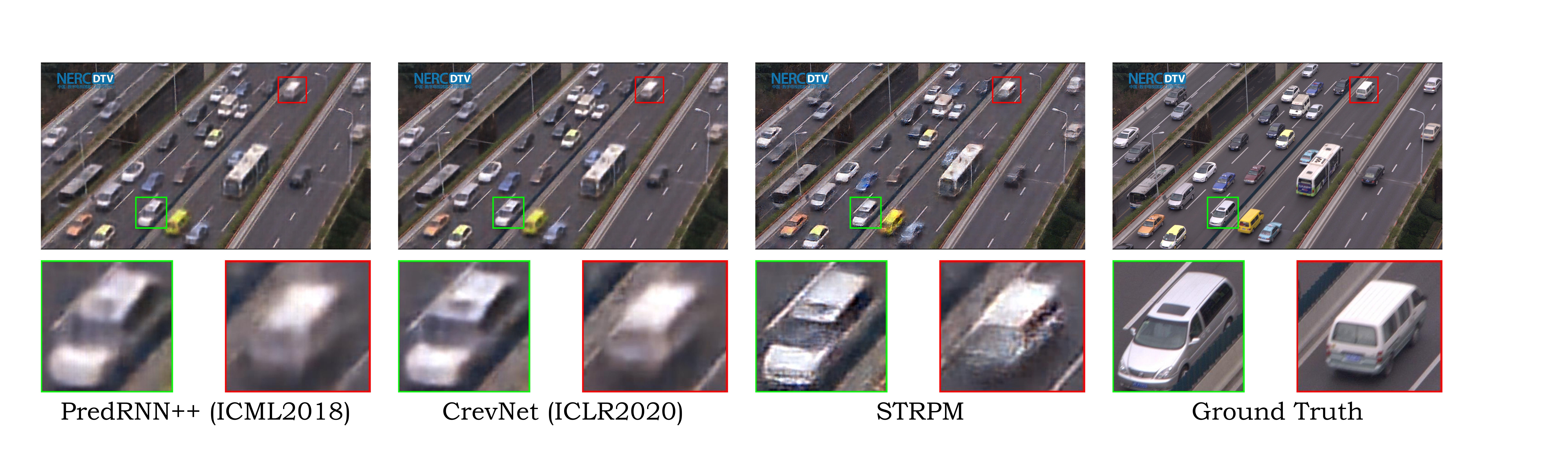}
  \caption{The generated examples on the SJTU4K dataset (4 frames $ \rightarrow$ 1 frame).}\label{fig:sjtu4k}
\end{figure*}
To further evaluate the efficiency of the proposed spatiotemporal encoding scheme and the residual predictive memory, we visualize the temporal features $T_E$, the spatial features $S_E$, the input information and the spatiotemporal residual information (STRF). The visualized results are shown in Figure \ref{fig:features}, where the temporal features contain a wider motion area while the spatial features focus on the appearance area of the person with greater weight values. The differences between the temporal features and the spatial features indicate that the spatiotemporal encoding scheme can help extract different features from the temporal domain and the spatial domain, respectively. Moreover, compared with the temporal features, the weights of the learned spatiotemporal residual features are much greater, indicating the proposed residual structure can help the predictive memory pay more attention to modeling the complex motion information instead of the appearance information (the weights of the appearance information is less than the spatial features).
\subsection{Human3.6M and SJTU4K Datasets}
The Human3.6M dataset consists of 3.6 million 3D human poses and corresponding images acted by 11 professional actors in 17 scenarios, including discussion, smoking, taking photos and so on. All videos are recorded with 4 calibrated cameras with a resolution $1000\times 1000$, which are further resized to $1024\times 1024$ in this paper. 73,404 sequences are for training and 8,582 for testing.
Figure \ref{fig:human36m} shows the generated examples from the proposed method and other state-of-the-art methods, where the proposed STRPM significantly outperforms others and the predicted results are more naturalistic.
In Table \ref{tab:ucf_human36m}, the proposed method achieves the best PSNR and LPIPS scores compared with other state-of-the-art methods.


The SJTU4K dataset consists of 15 ultra-high resolution 4K videos with a wide variety of contents. The resolution for each video is $2160\times 3840$. To evaluate the performance on ultra-high resolution videos, the inputs and outputs are all 4K videos without being down-sampled. 3,873 sequences are for training and 445 for testing.
To the best of our knowledge, the proposed STRPM is the first one predicting 4K videos. Figure \ref{fig:sjtu4k} shows the predicted 4K video frames from different methods.
The quantitative results are summarized in Table \ref{tab:sjtu4k}. As shown in Figure \ref{fig:sjtu4k} and Table \ref{tab:sjtu4k}, the proposed method has achieved the best qualitative and quantitative results on ultra-high-resolution videos with a satisfactory inference speed.
\begin{table}[!htb]
  \centering
  \setlength\tabcolsep{0.2pt}
  \caption{Quantitative results of different methods on the SJTU4K test set (4 frames $ \rightarrow$ 4 frames). The inference time over 10 samples have also been summarized.}
  \label{tab:sjtu4k}
   { \begin{tabular}{lccc}
    \toprule
    \multirow{2}{*}{Method}
    &$t=5$&$t=8$&Inference\cr
    \cmidrule(lr){2-2} \cmidrule(lr){3-3}
    &PSNR$\uparrow$/LPIPS$\downarrow$&PSNR$\uparrow$/LPIPS$\downarrow$&Time\cr
    \midrule
    ConvLSTM \cite{xingjian2015convolutional}              &22.74 / 67.81  &17.91 / 86.84&39.38s\cr
    PredRNN \cite{wang2017predrnn}                          &23.25 / 66.60  &18.20 / 87.04&40.06s\cr
    PredRNN++ \cite{wang2018predrnn++}                      &23.43 / 64.07  &18.55 / 86.34&53.11s\cr
    SAVP \cite{lee2018stochastic}                          &23.41 / 61.44  &18.63 / 80.45&100.23s\cr
    CrevNet \cite{yu2019efficient}                          &24.35 / 62.31  &19.61 / 80.91&52.98s\cr
    MotionRNN                                               &23.47 / 65.21 &19.72 / 81.39&61.87s\cr
    \midrule
    STRPM                                                   &\textbf{24.37} / \textbf{57.12}  &\textbf{19.77} / \textbf{66.68}&39.84s  \cr
    \bottomrule
    \end{tabular}}
\end{table}
\begin{table}[htb]
  \centering
\setlength\tabcolsep{1.5pt}
  \caption{Ablation studies on the proposed residual predictive memory and the spatiotemporal encoding-decoding scheme (\textbf{STED}) on the Human3.6M dataset (4 frames $\rightarrow$ 4 frames). PSNR and LPIPS scores are averaged over all 4 predictions.
  For a fair comparison, the encoders and decoders for all models are with the same structure and the number of the hidden state channels for all the memories is set to 128. We stack 16 memories into each model. All models are trained with MSE loss functions. The floating point operations (FLOPs) are recorded over 1 sample.}
  \label{tab:structure}
   { \begin{tabular}{lcccc}
    \toprule
    Method
    &PSNR$\uparrow$&LPIPS$\downarrow$&Parameters&FLOPs\cr
    \midrule
    Casual-LSTM \cite{wang2018predrnn++}                        &29.61  &14.42&131.35M&36.19G\cr
    E3D-LSTM \cite{wang2019eidetic}                             &29.93  &13.23&501.37M&128.82G\cr
    Reversible-PM \cite{yu2019efficient}                        &30.13  &12.62&114.35M&35.07G\cr
    \midrule
    RPM w/o residual                                            &30.11  &12.64&108.29M&30.55G\cr
    RPM ($\theta=1, \tau=1$)                                                &30.14  &12.58&\textbf{45.95M}&\textbf{14.96G}\cr
    RPM ($\theta=5, \tau=1$)                                                  &30.56  &11.98&77.09M&22.75G\cr
    RPM ($\theta=1, \tau=5$)                                                   &30.32  &12.31&77.09M&22.75G\cr
    RPM                                                         &31.10  &11.89&108.29M&30.55G\cr
    RPM + STED                                                 &\textbf{31.81}&\textbf{11.72}&109.13M&34.24G\\
    \bottomrule
    \end{tabular}}
\end{table}
\subsection{Ablation Study}
In this section, a series of ablation studies are conducted. Table \ref{tab:structure} shows the results of different models with different structures. For a fair comparison, all models without STED (the spatiotemporal encoding-decoding scheme) are built with the same structure except the predictive memory. Experimental results show that the proposed residual predictive memory outperforms other state-of-the-art memories with the lowest computation load and fewest parameters. In addition, the residual structure, the broadened temporal receptive field ($\tau>1$), and the broadened spatial receptive field ($\theta>1$) can help improve the performance of RPM. Moreover, RPM with STED also help improves the model performance.
\begin{table}[t]
  \centering
  \setlength\tabcolsep{3pt}
  \caption{Ablation studies on STRPM with different loss functions. Performance scores are averaged over all predictions.}
  \label{tab:loss}
   { \begin{tabular}{lcccc}
    \toprule
    \multirow{3}{*}{Method}
    &\multicolumn{2}{c}{UCF sports}&\multicolumn{2}{c}{Human3.6M}\cr
    &\multicolumn{2}{c}{$4\rightarrow6$}&\multicolumn{2}{c}{$4\rightarrow4$}\cr
    \cmidrule(lr){2-3} \cmidrule(lr){4-5}
    &PSNR$\uparrow$&LPIPS$\downarrow$&PSNR$\uparrow$&LPIPS$\downarrow$\cr
    \midrule
    $\mathcal{L}_{MSE}$                                                   &\textbf{24.89}  &38.81 &\textbf{31.81}  &11.72\cr
    $\mathcal{L}_{MSE}$+$\mathcal{L}_{GAN}$                               &23.98  &35.01  &30.53  &10.98\cr
    $\mathcal{L}_{MSE}$+$\mathcal{L}_{GAN}$+$\mathcal{L}_{LP}$            &24.30  &\textbf{31.50}  &31.00  &\textbf{10.11}\cr
    \bottomrule
    \end{tabular}}
\end{table}
Furthermore, results from methods trained with different loss functions are summarized in Table \ref{tab:loss}, where the proposed perceptual loss can help obtain a better trade-off between the objective quality (PSNR) and the perceptual quality (LPIPS).
\section{Conclusion and Discussions}
We proposed a Spatiotemporal Residual Predictive Model (STRPM) for High-Resolution Video Prediction.
We designed the Spatiotemporal Encoding-Decoding Scheme and the Residual Predictive Memory (RPM) to model the much more complex appearance information and motion information in high-resolution videos.
In addition, we proposed a Learned Perceptual loss to generate more naturalistic frames compared with the standard MSE loss.
Experimental results showed that the proposed model can predict high-resolution videos with the best objective and subjective quality compared with various existing methods.


Although the model performance is better than current methods, the practicality is still far from satisfactory, especially for videos with ultra-high resolutions ($\geq$1080p). In addition, the model efficiency also needs to be improved for multi-step predictions. Considering the above limitations, current predictive models maybe not reliable enough to be applied into decision-making systems that require data with high accuracy and real-time interactions, such as autonomous driving, robot control, etc. Further works are highly-encouraged to solve the above potential problems.
{\small
\bibliographystyle{ieee_fullname}
\bibliography{egbib}

\begin{thebibliography}{10}\itemsep=-1pt

\bibitem{babaeizadeh2018stochastic}
Mohammad Babaeizadeh, Chelsea Finn, Dumitru Erhan, Roy~H Campbell, and Sergey
  Levine.
\newblock Stochastic variational video prediction.
\newblock In {\em Int. Conf. Learn. Represent.}, 2018.

\bibitem{bhattacharyya2018long}
Apratim Bhattacharyya, Mario Fritz, and Bernt Schiele.
\newblock Long-term on-board prediction of people in traffic scenes under
  uncertainty.
\newblock In {\em IEEE Conf. Comput. Vis. Pattern Recog.}, pages 4194--4202,
  2018.

\bibitem{chang2021stae}
Zheng Chang, Xinfeng Zhang, Shanshe Wang, Siwei Ma, Yan Ye, and Wen Gao.
\newblock Stae: A spatiotemporal auto-encoder for high-resolution video
  prediction.
\newblock In {\em Int. Conf. Multimedia and Expo}, pages 1--6. IEEE, 2021.

\bibitem{chen2020long}
Xinyuan Chen, Chang Xu, Xiaokang Yang, and Dacheng Tao.
\newblock Long-term video prediction via criticization and retrospection.
\newblock {\em IEEE Trans. Image Process.}, 29:7090--7103, 2020.

\bibitem{denton2018stochastic}
Emily Denton and Rob Fergus.
\newblock Stochastic video generation with a learned prior.
\newblock In {\em Int. Conf. Mach. Learn.}, pages 1174--1183, 2018.

\bibitem{finn2016unsupervised}
Chelsea Finn, Ian Goodfellow, and Sergey Levine.
\newblock Unsupervised learning for physical interaction through video
  prediction.
\newblock In {\em Adv. Neural Inform. Process. Syst.}, pages 64--72, 2016.

\bibitem{franceschi2020stochastic}
Jean-Yves Franceschi, Edouard Delasalles, Micka{\"e}l Chen, Sylvain Lamprier,
  and Patrick Gallinari.
\newblock Stochastic latent residual video prediction.
\newblock In {\em Int. Conf. Mach. Learn.}, pages 3233--3246. PMLR, 2020.

\bibitem{goodfellow2014generative}
Ian Goodfellow, Jean Pouget-Abadie, Mehdi Mirza, Bing Xu, David Warde-Farley,
  Sherjil Ozair, Aaron Courville, and Yoshua Bengio.
\newblock Generative adversarial nets.
\newblock In {\em Adv. Neural Inform. Process. Syst.}, pages 2672--2680, 2014.

\bibitem{hochreiter1997long}
Sepp Hochreiter and J{\"u}rgen Schmidhuber.
\newblock Long short-term memory.
\newblock {\em Neural Computation}, 9(8):1735--1780, 1997.

\bibitem{h36m_pami}
Catalin Ionescu, Dragos Papava, Vlad Olaru, and Cristian Sminchisescu.
\newblock Human3.6m: Large scale datasets and predictive methods for 3d human
  sensing in natural environments.
\newblock {\em IEEE Trans. Pattern Anal. Mach. Intell.}, 2014.

\bibitem{jin2020exploring}
Beibei Jin, Yu Hu, Qiankun Tang, Jingyu Niu, Zhiping Shi, Yinhe Han, and
  Xiaowei Li.
\newblock Exploring spatial-temporal multi-frequency analysis for high-fidelity
  and temporal-consistency video prediction.
\newblock In {\em IEEE Conf. Comput. Vis. Pattern Recog.}, pages 4554--4563,
  2020.

\bibitem{kwon2019predicting}
Yong-Hoon Kwon and Min-Gyu Park.
\newblock Predicting future frames using retrospective cycle gan.
\newblock In {\em IEEE Conf. Comput. Vis. Pattern Recog.}, pages 1811--1820,
  2019.

\bibitem{lee2018stochastic}
Alex~X Lee, Richard Zhang, Frederik Ebert, Pieter Abbeel, Chelsea Finn, and
  Sergey Levine.
\newblock Stochastic adversarial video prediction.
\newblock {\em arXiv preprint arXiv:1804.01523}, 2018.

\bibitem{ma2019image}
Siwei Ma, Xinfeng Zhang, Chuanmin Jia, Zhenghui Zhao, Shiqi Wang, and Shanshe
  Wang.
\newblock Image and video compression with neural networks: A review.
\newblock {\em IEEE Trans. Circuit Syst. Video Technol.}, 2019.

\bibitem{mathieu2016deep}
Michael Mathieu, Camille Couprie, and Yann LeCun.
\newblock Deep multi-scale video prediction beyond mean square error.
\newblock In {\em Int. Conf. Learn. Represent.}, 2016.

\bibitem{meyer2018phasenet}
Simone Meyer, Abdelaziz Djelouah, Brian McWilliams, Alexander Sorkine-Hornung,
  Markus Gross, and Christopher Schroers.
\newblock Phasenet for video frame interpolation.
\newblock In {\em IEEE Conf. Comput. Vis. Pattern Recog.}, pages 498--507,
  2018.

\bibitem{niklaus2018context}
Simon Niklaus and Feng Liu.
\newblock Context-aware synthesis for video frame interpolation.
\newblock In {\em IEEE Conf. Comput. Vis. Pattern Recog.}, pages 1701--1710,
  2018.

\bibitem{ranzato2014video}
MarcAurelio Ranzato, Arthur Szlam, Joan Bruna, Michael Mathieu, Ronan
  Collobert, and Sumit Chopra.
\newblock Video (language) modeling: a baseline for generative models of
  natural videos.
\newblock {\em arXiv preprint arXiv:1412.6604}, 2014.

\bibitem{rodriguez2008action}
Mikel~D Rodriguez, Javed Ahmed, and Mubarak Shah.
\newblock Action mach a spatio-temporal maximum average correlation height
  filter for action recognition.
\newblock In {\em IEEE Conf. Comput. Vis. Pattern Recog.}, pages 1--8, 2008.

\bibitem{xingjian2015convolutional}
Xingjian Shi, Zhourong Chen, Hao Wang, Dit-Yan Yeung, Wai-Kin Wong, and
  Wang-chun Woo.
\newblock Convolutional lstm network: A machine learning approach for
  precipitation nowcasting.
\newblock In {\em Adv. Neural Inform. Process. Syst.}, pages 802--810, 2015.

\bibitem{song2013sjtu}
Li Song, Xun Tang, Wei Zhang, Xiaokang Yang, and Pingjian Xia.
\newblock The sjtu 4k video sequence dataset.
\newblock In {\em International Workshop on Quality of Multimedia Experience},
  pages 34--35. IEEE, 2013.

\bibitem{srivastava2015unsupervised}
Nitish Srivastava, Elman Mansimov, and Ruslan Salakhudinov.
\newblock Unsupervised learning of video representations using lstms.
\newblock In {\em Int. Conf. Mach. Learn.}, pages 843--852, 2015.

\bibitem{villegas2019high}
Ruben Villegas, Arkanath Pathak, Harini Kannan, Dumitru Erhan, Quoc~V Le, and
  Honglak Lee.
\newblock High fidelity video prediction with large stochastic recurrent neural
  networks.
\newblock In {\em Adv. Neural Inform. Process. Syst.}, 2019.

\bibitem{wang2018predrnn++}
Yunbo Wang, Zhifeng Gao, Mingsheng Long, Jianmin Wang, and S~Yu Philip.
\newblock Predrnn++: Towards a resolution of the deep-in-time dilemma in
  spatiotemporal predictive learning.
\newblock In {\em Int. Conf. Mach. Learn.}, pages 5123--5132, 2018.

\bibitem{wang2019eidetic}
Yunbo Wang, Lu Jiang, Ming-Hsuan Yang, Li-Jia Li, Mingsheng Long, and Li.
  Fei-Fei.
\newblock Eidetic 3d lstm: A model for video prediction and beyond.
\newblock In {\em Int. Conf. Learn. Represent.}, 2019.

\bibitem{wang2017predrnn}
Yunbo Wang, Mingsheng Long, Jianmin Wang, Zhifeng Gao, and S~Yu Philip.
\newblock Predrnn: Recurrent neural networks for predictive learning using
  spatiotemporal lstms.
\newblock In {\em Adv. Neural Inform. Process. Syst.}, pages 879--888, 2017.

\bibitem{wu2021motionrnn}
Haixu Wu, Zhiyu Yao, Jianmin Wang, and Mingsheng Long.
\newblock Motionrnn: A flexible model for video prediction with
  spacetime-varying motions.
\newblock In {\em IEEE Conf. Comput. Vis. Pattern Recog.}, pages 15435--15444,
  2021.

\bibitem{xu2020video}
Jingwei Xu, Huazhe Xu, Bingbing Ni, Xiaokang Yang, and Trevor Darrell.
\newblock Video prediction via example guidance.
\newblock In {\em Int. Conf. Mach. Learn.}, pages 10628--10637. PMLR, 2020.

\bibitem{yu2019efficient}
Wei Yu, Yichao Lu, Steve Easterbrook, and Sanja Fidler.
\newblock Efficient and information-preserving future frame prediction and
  beyond.
\newblock In {\em Int. Conf. Learn. Represent.}, 2020.

\bibitem{zhang2018unreasonable}
Richard Zhang, Phillip Isola, Alexei~A Efros, Eli Shechtman, and Oliver Wang.
\newblock The unreasonable effectiveness of deep features as a perceptual
  metric.
\newblock In {\em IEEE Conf. Comput. Vis. Pattern Recog.}, pages 586--595,
  2018.

\bibitem{zhu2017unpaired}
Jun-Yan Zhu, Taesung Park, Phillip Isola, and Alexei~A Efros.
\newblock Unpaired image-to-image translation using cycle-consistent
  adversarial networks.
\newblock In {\em Int. Conf. Comput. Vis.}, pages 2223--2232, 2017.

\end{thebibliography}
}

\end{document}